\documentclass[11pt]{article}

\usepackage[preprint]{acl}

\usepackage{times}
\usepackage{latexsym}
\usepackage{booktabs}
\usepackage{enumitem}
\usepackage{graphicx}  
\usepackage{tcolorbox}
\tcbuselibrary{breakable}
\usepackage[T1]{fontenc}
\usepackage{algorithm}
\usepackage{algorithmic}
\usepackage[utf8]{inputenc}
\usepackage{amsmath}
\usepackage{microtype}
\usepackage{amssymb}
\usepackage{inconsolata}

\usepackage{graphicx}

\title{AMR-SD: Asymmetric Meta-Reflective Self-Distillation for Token-Level Credit Assignment}

\author{
    \textbf{Zhenlin Wei\textsuperscript{1,2}\footnotemark[1]},
    \textbf{Pu Jian\textsuperscript{2}\footnotemark[1]},
    \textbf{Yingzhuo Deng\textsuperscript{2}\footnotemark[1]},
    \textbf{Xiaohan Wang\textsuperscript{1}\footnotemark[2]},
    \textbf{Jiajun Chai\textsuperscript{1}},
    \textbf{Zhexin Hu\textsuperscript{2}},\\
    \textbf{Wei Lin\textsuperscript{1}},
    \textbf{Shanbin Zhang\textsuperscript{2}},
    \textbf{Guojun Yin\textsuperscript{1}}
    \\
    \textbf{\textsuperscript{1}} Meituan Beijing, China \\
    \textbf{\textsuperscript{2}} University of Chinese Academy of Sciences \\
    \texttt{\small \ wangxiaohan17@meituan.com}
}

\begin{document}
\maketitle
\renewcommand{\thefootnote}{\fnsymbol{footnote}}
\footnotetext[1]{\ \ Equal contribution.}
\footnotetext[2]{\ \ Corresponding author.}
\footnotetext[3]{\ \ This work was done during an internship at Meituan, supervised by Xiaohan Wang.}
\renewcommand{\thefootnote}{\arabic{footnote}}
\begin{abstract}
The alignment of Large Language Models (LLMs) for complex reasoning heavily relies on Reinforcement Learning with Verifiable Rewards (RLVR). However, standard algorithms like GRPO apply sequence-level rewards uniformly to all tokens, creating a severe credit-assignment bottleneck. While on-policy self-distillation attempts to resolve this by conditioning a self-teacher on privileged contexts, direct exposure to raw oracle solutions often induces over-conditioned teacher distributions, implicit answer leakage, and late-stage training collapse. To overcome these limitations, we propose Asymmetric Meta-Reflective Self-Distillation (AMR-SD). Instead of conditioning directly on raw reference traces, AMR-SD inserts a reflection bottleneck: it compresses diagnostic signals—from verifier outcomes, peer rollouts, or reference feedback—into concise, self-generated Socratic hints and critiques. Furthermore, we introduce Causal Information Gain (CIG) with an asymmetric, ReLU-gated threshold to translate these reflections into sparse, highly precise token-level advantage modulations. Combined with temporal annealing, this mechanism preserves the base environmental reward while filtering out distributional noise. Experiments across scientific, mathematical, and tool-use benchmarks demonstrate that AMR-SD significantly outperforms existing baselines, achieving robust long-horizon stability and successfully preventing late-stage collapse.
\end{abstract}

\begin{figure}[t]
  \centering
  \includegraphics[width=1\columnwidth]{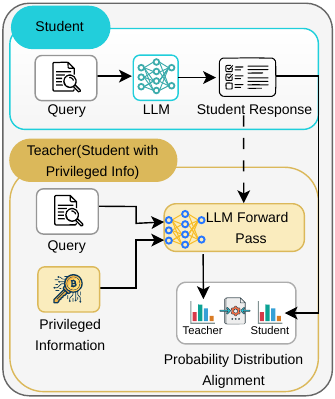}

  \caption{In standard on-policy self-distillation, the student generates a response, which a teacher evaluates via forced-decoding with privileged information to produce token-level probabilities for student alignment.}
  \label{intro:case}
\end{figure}

\section{Introduction}

The alignment of Large Language Models (LLMs) for complex reasoning heavily relies on Reinforcement Learning with Verifiable Rewards (RLVR)~\cite{yu2025rlpr,wu2025invisible}. Standard policy gradient algorithms, such as Group Relative Policy Optimization (GRPO)~\cite{guo2025deepseek,schulman2017proximal}, efficiently estimate advantages from sparse sequence-level outcomes. However, GRPO applies a uniform advantage scalar to all tokens within a rollout. This creates a severe credit-assignment bottleneck: critical logical deductions and superficial stylistic choices receive identical reward signals, hampering the optimization of reasoning tasks~\cite{yang2026your,li2026outcome,xie2025unlocking,li2026rubrichub}.

To resolve this credit-assignment bottleneck, offline Supervised Fine-Tuning (SFT) offers dense supervision but suffers from severe train-test distribution shifts. To bridge the gap between RL's on-policy exploration and SFT's dense supervision, the community turned to On-Policy Distillation (OPD)~\cite{agarwal2024policy}. While highly effective~\cite{xiao2026mimo,team2026kimi}, maintaining an external teacher incurs prohibitive computational overhead. Consequently, On-Policy Self-Distillation emerged, utilizing the model itself as the teacher conditioned on privileged information (e.g., ground-truth solutions) (Figure~\ref{intro:case})~\cite{zhao2026self,hubotter2026reinforcement,shenfeld2026self,zhang2026embarrassingly}. However, directly forcing a conditionally independent student to match a conditionally dependent teacher introduces information asymmetry. The student blindly imitates confident reasoning without accessing the underlying context, leading to privileged information leakage, degraded intrinsic reflective capabilities, and late-stage training collapse~\cite{kim2026does,li2026rethinking}.

Recent advancements attempt to mitigate this by anchoring optimization to environment rewards for continuous magnitude adjustment~\cite{yang2026self,li2026unifying}. Unfortunately, a subtle limitation remains: because the privileged context narrows the teacher's output distribution onto specific oracle tokens, the teacher's probabilities for most student-generated tokens are systematically lower than the student's own. This causes the continuous multipliers to be predominantly less than 1, dampening the primary reward signal. Furthermore, the privileged information here is often simply the directly provided ground truth; directly relying on it during training is clearly inefficient. Consequently, such global distribution alignment inevitably allows privileged information to implicitly leak into the parameters, inducing late-stage training collapse.

To fully unlock self-distillation without these drawbacks, we propose the Asymmetric Meta-Reflective Self-Distillation (AMR-SD) framework, which retains the reward-anchoring principle of~\cite{yang2026self} while replacing continuous 
scaling with an asymmetric, ReLU-gated threshold ($\tau$), powered by a novel metric: Causal Information Gain (CIG). To mitigate direct oracle-solution leakage, AMR-SD operates as a reflection-mediated self-training loop. Instead of exposing the teacher to raw ground-truth solutions during token rescoring, we enforce a Meta-Reflection phase: the model generates an encouraging \texttt{<hint>} for successful trajectories or a targeted \texttt{<critique>} for failed trajectories, using verifier-approved peer rollouts when available, which then serve as the conditioning context for the teacher.

Crucially, to overcome the signal dampening inherent in continuous scaling, CIG introduces an asymmetric, threshold-gated filtering mechanism. Rather than diluting the reward by modulating every token, this mechanism acts as a strict gatekeeper: it preserves the baseline environmental reward intact by default, triggering sparse, targeted adjustments only when the Socratic teacher identifies a highly confident divergence from the student. This approach effectively filters out trivial distributional noise, delivering precise token-level credit assignment without compromising the overall strength of the base reinforcement signal.

We emphasize that AMR-SD does not entirely discard privileged information (e.g., verifier outcomes, peer rollouts, or reference answers). Instead, it prevents the token-rescoring teacher from directly seeing raw oracle traces. Diagnostic sources are first compressed into concise natural-language reflections, acting as a low-bandwidth bottleneck. This mitigates over-conditioned teacher distributions and answer leakage while preserving essential token-level guidance for credit assignment.

Our main contributions are threefold:
\begin{itemize}[leftmargin=15pt]
\item We introduce Meta-Reflection, prompting the model to generate Socratic self-teaching targets (\texttt{hint} / \texttt{critique}) from verifier outcomes, on-policy peer rollouts, or reference-assisted diagnostic signals when available. This reflection-mediated paradigm does not remove privileged information, but introduces a low-bandwidth bottleneck between diagnostic sources and token-level rescoring, thereby reducing direct oracle-trace conditioning and improving calibration of token-level credit signals.

\item We design CIG as a threshold-gated (ReLU) advantage adjustment mechanism. Integrated with temporal annealing, it concentrates dense supervision during early learning and smoothly shifts to unadulterated environmental rewards, ensuring rapid early alignment and long-horizon stability.
\item Building upon these mechanisms, we propose the AMR-SD framework. Extensive experiments across scientific, mathematical, and tool-use benchmarks demonstrate that AMR-SD outperforms existing baselines, achieving robust long-horizon stability and successfully preventing the late-stage training collapse.
\end{itemize}

\section{Related Work and Background}

\paragraph{RL with Verifiable Rewards and GRPO.}
Proximal Policy Optimization~\cite{schulman2017proximal} established the foundation for RLVR-based LLM alignment~\cite{yu2025rlpr,wu2025invisible}. Recent works have further analyzed the properties of verifiable reward signals~\cite{wen2025reinforcement,lu2026contextual,lin2026resrl}. To optimize the policy under sparse rewards without a separate value network, GRPO~\cite{guo2025deepseek} estimates advantages via group baseline normalization. For a given prompt $x$, the policy samples $G$ independent trajectories. The advantage for the $i$-th trajectory is calculated as $A_i = (r_i - \mu_{\text{group}})/(\sigma_{\text{group}} + \varepsilon)$, where $\mu_{\text{group}}$ and $\sigma_{\text{group}}$ are the mean and standard deviation of the $G$ rewards. 

Although subsequent systems such as DAPO~\cite{yu2026dapo} improve training stability, GRPO's advantage estimates are known to be biased~\cite{yang2026your}. More critically, applying a uniform scalar advantage $A_i$ to every token creates a severe credit-assignment bottleneck~\cite{yang2026your,li2026outcome,xie2025unlocking,li2026rubrichub}. Some works attempt to address this through uncertainty-aware advantage shaping~\cite{xie2025unlocking}, outcome-grounded advantage reshaping~\cite{li2026outcome}, and token-level policy gradient redesign for tool use~\cite{lin2025rest,lin2025awpo,lin2026resrl}, yet fine-grained credit assignment remains challenging.

\paragraph{On-Policy (Self-)Distillation.}
To provide denser, token-level supervision, On-Policy Distillation (OPD)~\cite{agarwal2024policy} trains a student on its own online rollouts evaluated by a stronger external teacher (e.g., MiMo~\cite{xiao2026mimo}, Kimi~\cite{team2026kimi}). However, maintaining an external teacher imposes substantial computational overhead. To eliminate this dependency, On-Policy Self-Distillation~\cite{zhao2026self,hubotter2026reinforcement,shenfeld2026self,zhang2026embarrassingly} uses the model itself as the teacher. In this paradigm, the active student policy $\pi_\theta$ generates trajectories, and the exact same model, detached via a stop-gradient operator ($\pi_{\theta_{\text{sg}}}$), acts as the teacher conditioned on privileged information $c$. 

Despite its efficiency, prior work~\cite{kim2026does,li2026rethinking} identifies that directly aligning the student to these conditional log-likelihoods introduces privileged information leakage and late-stage collapse. Recent methods~\cite{yang2026self,li2026unifying} attempt to stabilize training by anchoring to environment rewards and using the teacher only for continuous magnitude adjustment. Yet, their continuous scaling predominantly dampens the overall learning signal. Our AMR-SD framework resolves these issues via autonomous Meta-Reflection and a ReLU-gated threshold, achieving sparse, unbiased token-level credit assignment without external dependencies.

\begin{figure*}[t]
\centering
\begin{minipage}{\textwidth}
\begin{algorithm}[H]
\caption{CIG-Enhanced AMR-SD Training}
\label{alg:amr_sd}
\textbf{Input:} Active policy $\pi_\theta$, Dataset $\mathcal{D}$;
Hyperparameters: $\lambda, \gamma \in \mathbb{R}_{>0}$, $\tau \geq 0$, $\kappa > 0$, $T_{\text{decay}} \in \mathbb{N}$
\begin{algorithmic}[1]
\FOR{each iteration $t_{\text{global}}$}
    \STATE $\theta_{\text{old}} \leftarrow \text{stop\_gradient}(\theta)$; $\theta_{\text{sg}} \leftarrow \theta_{\text{old}}$
    \hfill \textit{\small // Frozen old policy and stop-gradient teacher}
    \STATE $\{\lambda_{\text{eff}}, \gamma_{\text{eff}}\} \leftarrow \{\lambda, \gamma\} \cdot \max(0, 1 - t_{\text{global}}/T_{\text{decay}})$
    \hfill \textit{\small // Temporal annealing}
    \FOR{each mini-batch $\mathcal{B} \sim \mathcal{D}$}
        \FOR{each prompt $x \in \mathcal{B}$}
            \STATE Sample $G$ trajectories $\{y_i\}_{i=1}^{G} \sim \pi_{\theta_{\text{old}}}(\cdot \mid x)$; evaluate rewards $\{r_i\}_{i=1}^{G}$
            \STATE Compute normalized advantages $A_i = ({r_i - \mu_{\text{group}}})({\sigma_{\text{group}} + \varepsilon})^{-1}$
            \STATE $\mathcal{P}_x \leftarrow \{y_j \mid r_j = 1,\ y_j \in \{y_i\}_{i=1}^{G}\}$
            \hfill \textit{\small // Same-group verifier-approved peers}
            \FOR{each trajectory $y_i$}
                \IF{$A_i \geq 0$}
                    \STATE $c_i \leftarrow \text{GenerateHint}_{\theta_{\text{sg}}}(x, y_i)$; $m_i \leftarrow 1$
                \ELSIF{$\mathcal{P}_x \neq \emptyset$}
                    \STATE Select $y_i^+ \in \mathcal{P}_x$; 
                    $c_i \leftarrow \text{GenerateCritique}_{\theta_{\text{sg}}}(x, y_i, y_i^+)$; $m_i \leftarrow 1$
                \ELSE
                    \STATE $c_i \leftarrow \emptyset$; $m_i \leftarrow 0$
                    \hfill \textit{\small // Fallback to GRPO}
                \ENDIF
                \STATE Obtain $\log \pi_{\theta_{\text{sg}}}(y_{i,t} \mid x, c_i, y_{i,<t})$ and $\log \pi_\theta(y_{i,t} \mid x, y_{i,<t})$ for all $t$ via forced decoding
                \hfill \textit{\small // Teacher conditional and Student unconditional scores}
                \FOR{$t = 1$ to $T_i$}
                    \STATE $\widehat{\mathcal{I}}_{i,t}^{\text{CIG}} \leftarrow \operatorname{clip}\!\left(\log \pi_{\theta_{\text{sg}}}(y_{i,t} \mid x, c_i, y_{i,<t}) - \log \pi_\theta(y_{i,t} \mid x, y_{i,<t}),\ -\kappa,\ \kappa\right)$
                    \STATE $\Delta_{i,t} \leftarrow m_i \cdot \left[
                    \mathbb{I}_{\{A_i \geq 0\}}\lambda_{\text{eff}}\max(0,\widehat{\mathcal{I}}_{i,t}^{\text{CIG}}-\tau)
                    +
                    \mathbb{I}_{\{A_i < 0\}}\gamma_{\text{eff}}\max(0,-\widehat{\mathcal{I}}_{i,t}^{\text{CIG}}-\tau)
                    \right]$
                    \STATE $\widehat{A}_{i,t} \leftarrow \operatorname{stop\_gradient}\!\left(A_i \cdot (1 + \Delta_{i,t})\right)$
                \ENDFOR
            \ENDFOR
        \ENDFOR
        \STATE Update $\theta$ by maximizing $\mathcal{L}_{\text{AMR-SD}}(\theta) = \mathbb{E}_{i \sim \mathcal{B}}\!\left[\frac{1}{T_i}\sum_{t=1}^{T_i} \min\!\left(\rho_{i,t}\widehat{A}_{i,t},\ \operatorname{clip}(\rho_{i,t}, 1-\varepsilon, 1+\varepsilon)\widehat{A}_{i,t}\right)\right]$
    \ENDFOR
\ENDFOR
\end{algorithmic}
\end{algorithm}
\end{minipage}
\end{figure*}

\section{Method}

We introduce the AMR-SD framework. Unlike standard sequence-level RL which broadcasts a uniform advantage to all tokens, AMR-SD resolves the spatial credit-assignment bottleneck by dynamically mapping the sequence-level GRPO advantage into a dense, state-action specific tensor. 

Formally, we model the language generation process as a MDP defined by the tuple $(\mathcal{S}, \mathcal{A}, \mathcal{P}, \mathcal{R})$. Let the state $s_t = (x, y_{<t}) \in \mathcal{S}$ denote the prompt and the generated prefix, and the action $a_t = y_t \in \mathcal{A}$ represent the current token.  To construct a precise token-level advantage $\widehat{A}_{i,t}$ without suffering from the information leakage inherent in traditional self-distillation, our framework operates in three sequential phases: Socratic Rescoring via Meta-Reflection, CIG quantification, and Asymmetric Modulation.The full procedure is summarized in Algorithm~\ref{alg:amr_sd}.

\subsection{Socratic Rescoring via Meta-Reflection}

To construct a dense reward signal without human annotation, we formalize a Meta-Reflection mechanism that transitions the stop-gradient policy $\pi_{\theta_{\text{sg}}}$ into a conditional Socratic evaluator. Let $\mathcal{C}$ be the latent space of natural language reflections. We define a discrete conditional mapping function $f_{\text{reflect}}: \mathcal{S} \times \mathcal{A}^T \times \mathbb{R} \to \mathcal{C}$ governed by the signum of the empirical advantage:
\begin{equation}
\begin{split}
    c_i &= f_{\text{reflect}}(x, y_i, \text{sgn}(A_i)) \\
    &= \mathbb{I}_{\{A_i \geq 0\}} \cdot c_{\text{hint}} + \mathbb{I}_{\{A_i < 0\}} \cdot c_{\text{critique}}
\end{split}
\end{equation}

where $\mathbb{I}_{\{\cdot\}}$ denotes the indicator function. The hint reflection $c_{\text{hint}}$ extracts latent reasoning trajectories for successful rollouts, while the critique reflection $c_{\text{critique}}$ isolates logical or syntactic divergences in failed ones. 

Crucially, the Meta-Reflection phase is executed in-situ by the current student policy $\pi_\theta$ itself. By generating hints and critiques as intermediate Socratic reflections, AMR-SD does not expose the rescoring teacher to raw oracle answers or full ground-truth reasoning traces. Instead, outcome information is first compressed into concise, structured natural-language guidance. This lower-bandwidth conditioning mitigates the over-conditioning and answer-leakage risks of directly conditioning the teacher on oracle solutions, while still providing token-level guidance for credit assignment.

The teacher model subsequently evaluates the trajectory via forced decoding over the augmented context vectors $[x \oplus c_i \oplus y_i]$. This operation yields a conditional state-transition probability distribution, defined as $\pi_{\theta_{\text{sg}}}(a_t \mid s_t, c_i)$. By injecting $c_i$, we effectively break the standard autoregressive Markov assumption, allowing the teacher to retrospectively assess action $a_t$ given the terminal trajectory outcome encoded in $c_i$.

\subsection{Causal Information Gain (CIG)}

To quantify the granular contribution of each action, we introduce CIG. From an information-theoretic perspective, CIG measures the pointwise log-likelihood ratio between the context-aware Socratic teacher and the student policy. For an action $a_t$ at state $s_t$, the raw CIG is computed as:
\begin{equation}
    \mathcal{I}_t^{\text{CIG}} = \log \left( \frac{\pi_{\theta_{\text{sg}}}(a_t \mid s_t, c_i)}{\pi_\theta(a_t \mid s_t)} \right)
\end{equation}

A strictly positive $\mathcal{I}_t^{\text{CIG}}$ indicates that the transition $(s_t, a_t)$ is assigned a low probability by the student but a high likelihood by the informed teacher, signaling a knowledge deficit. Conversely, $\mathcal{I}_t^{\text{CIG}} < 0$ implies student over-confidence relative to the Socratic baseline.

To guarantee bounded Lipschitz continuity in the subsequent gradient updates and mitigate numerical instability caused by the heavy-tailed nature of KL-divergence, we apply a symmetric clamping operator to the information gain:

\begin{equation}
    \widehat{\mathcal{I}}_t^{\text{CIG}} = \operatorname{clip}(\mathcal{I}_t^{\text{CIG}}, -\kappa, \kappa)
\end{equation}

where $\kappa \in \mathbb{R}^+$ defines the maximum allowed informational deviation. Based on our empirical observations of the CIG distribution (detailed in Appendix \ref{app:cig_distribution}), we constrain this bound to $\kappa = 5$ to mitigate numerical instability while preserving meaningful signals.

\subsection{Asymmetric Token-Level Advantage Modulation}

We integrate the localized CIG with the global trajectory objective through a non-linear, asymmetric modulation scalar $\Delta_t$. The asymmetry is mathematically parameterized to decouple the gradient flows for positive and negative rollouts, utilizing a noise-filtering threshold $\tau \in \mathbb{R}^+$:
\begin{equation}
\begin{split}
    \Delta_t = \;& \mathbb{I}_{\{A_i \geq 0\}} \cdot \lambda_{\text{eff}} \max(0, \widehat{\mathcal{I}}_t^{\text{CIG}} - \tau) \\
    &+ \mathbb{I}_{\{A_i < 0\}} \cdot \gamma_{\text{eff}} \max(0, -\widehat{\mathcal{I}}_t^{\text{CIG}} - \tau)
\end{split}
\end{equation}

This formulation ensures orthogonal gradient scaling:
\begin{itemize}
    \item \textbf{Reward Amplification:} When $A_i \geq 0$, $\Delta_t$ monotonically increases only for actions where the student underperformed the teacher ($\widehat{\mathcal{I}}_t^{\text{CIG}} > \tau$).
    \item \textbf{Penalty Amplification:} When $A_i < 0$, $\Delta_t$ monotonically increases only for actions where the student was erroneously over-confident ($\widehat{\mathcal{I}}_t^{\text{CIG}} < -\tau$).
\end{itemize}

The refined, token-level advantage tensor is thus expressed as a multiplicative modulation of the original baseline:
\begin{equation}
    \widehat{A}_{i,t} = A_i \cdot \left( 1 + \Delta_t \right)
\end{equation}

\paragraph{Temporal Annealing Schedule} 
In late training stages, self-distillation encounters two bottlenecks. First, reflection quality degrades: hints become redundant for internalized reasoning, and residual errors become too complex for accurate critiques. Second, the teacher reaches its intrinsic capability ceiling, losing the ability to provide meaningful CIG signals to an already highly-optimized student. To prevent overfitting to these depleted signals, a linear temporal decay smoothly shifts the optimization entirely to the pure environmental reward:
\begin{equation}
    \{\lambda_{\text{eff}}, \gamma_{\text{eff}}\} = \{\lambda, \gamma\} \cdot \max\left(0, 1 - \frac{t_{\text{global}}}{T_{\text{decay}}}\right)
\end{equation}
where $t_{\text{global}}$ is current optimization step.

\subsection{Optimization Objective}

The student policy parameters $\theta$ are optimized by maximizing a modified PPO clipped surrogate objective. We substitute the standard trajectory advantage with our dense estimator $\widehat{A}_{i,t}$:
\begin{equation}
     \mathcal{L}_{\text{AMR-SD}}(\theta) = \mathbb{E}_{x \sim \mathcal{D}, y \sim \pi_{\theta_{\text{old}}}} \Bigg[ \frac{1}{T_i} \sum_{t=1}^{T_i} \mathcal{L}_{\text{clip}}(\theta, t) \Bigg] 
\end{equation}

where the clipped surrogate term is defined as:
\begin{equation}
\begin{split}
    \mathcal{L}_{\text{clip}}(\theta, t) = \min \Big( &\rho_t(\theta) \widehat{A}_{i,t}, \\
    &\text{clip}(\rho_t(\theta), 1-\varepsilon, 1+\varepsilon) \widehat{A}_{i,t} \Big)
\end{split}
\end{equation}
and the importance sampling ratio is $\rho_t(\theta) = \frac{\pi_\theta(a_t \mid s_t)}{\pi_{\theta_{\text{old}}}(a_t \mid s_t)}$. The gradient $\nabla_\theta \mathcal{L}_{\text{AMR-SD}}(\theta)$ now incorporates both global trajectory optimality and precise token-level informational corrections.

\section{Experiment}

\subsection{Experimental Setup}
We evaluate AMR-SD on: (1) \textbf{Scientific Reasoning \& Tool Use} using SciKnowEval~\cite{feng2024sciknoweval} and ToolAlpaca~\cite{tang2023toolalpaca}; (2) \textbf{Mathematical Reasoning} using DAPO-Math-17k~\cite{yu2026dapo} and benchmarks AIME 24/25, AMC 23, and HMMT. Base models are \textbf{Qwen3-8B} and \textbf{Qwen2.5-7B-Instruct}~\cite{yang2025qwen3,bai2023qwen}. We compare against \textbf{GRPO}, \textbf{SDPO}, and \textbf{RLSD}. Models are trained via \textit{verl}~\cite{sheng2024hybridflow} ($G=8$, $T_{\text{decay}}=50$). All tracks are evaluated using \textit{acc@16}, with scientific and tool-use tasks reported at steps 15 and 75 to monitor stability. See Appendix \ref{app:hyper} for data splits, hyperparameters details.

\subsection{Main Result}
\begin{table*}[t]
  \centering
  \caption{Performance comparison on SciKnowEval and Tool Use. Accuracy at early and late training stages is reported to illustrate initial convergence speed versus long-horizon stability. Best results are in bold.}
  \label{tab:main_result}
  \begin{tabular}{l cccccccccc} 
\toprule
& \multicolumn{2}{c}{\textbf{Chemistry}}
& \multicolumn{2}{c}{\textbf{Physics}}
& \multicolumn{2}{c}{\textbf{Biology}}
& \multicolumn{2}{c}{\textbf{Materials}}
& \multicolumn{2}{c}{\textbf{Tool Use}} \\
\cmidrule(lr){2-3}\cmidrule(lr){4-5}\cmidrule(lr){6-7}\cmidrule(lr){8-9}\cmidrule(lr){10-11}
\textbf{Training Steps}& {15} & {75} & {15} & {75} & {15} & {75} & {15} & {75} & {15} & {75} \\
\midrule

\textbf{Qwen3-8B} & \multicolumn{2}{c}{40.5} & \multicolumn{2}{c}{57.4} & \multicolumn{2}{c}{28.7} & \multicolumn{2}{c}{59.1} & \multicolumn{2}{c}{57.3} \\
\ + GRPO~\cite{guo2025deepseek}
& 64.9 & 76.5 & 61.8 & 74.3 & 46.1 & 60.5 & 73.6 & \textbf{79.9} & 63.7 & 66.9 \\
\ + SDPO~\cite{hubotter2026reinforcement}
& \textbf{69.2} & 77.2 & 62.4 & 72.8 & 46.9 & 56.3 & 69.2 & 74.5 & 64.2 & - \\
\ + RLSD~\cite{yang2026self}
& 66.3 & \textbf{80.1} & 59.7 & 69.6 & \textbf{47.6} & 56.5 & 74.2 & 76.7 & \textbf{64.3} & 65.5 \\
\ + \textbf{AMR-SD (Ours)}
& 65.0 & 75.5 & \textbf{62.6} & \textbf{75.9} & 47.0 & \textbf{63.7} & \textbf{74.8} & 79.2 & 63.2 & \textbf{67.7} \\
\addlinespace[2pt]
\midrule

\textbf{Qwen2.5-7B-Instruct} & \multicolumn{2}{c}{28.6} & \multicolumn{2}{c}{55.6} & \multicolumn{2}{c}{29.0} & \multicolumn{2}{c}{57.5} & \multicolumn{2}{c}{52.4} \\
\ + GRPO~\cite{guo2025deepseek}
& 59.4 & 77.5 & 65.0 & 77.7 & 45.0 & 48.2 & 70.9 & 74.2 & 61.6 & \textbf{69.2} \\
\ + SDPO~\cite{hubotter2026reinforcement}
& 59.5 & 74.1 & 59.1 & 68.2 & 44.5 & 51.8 & 59.1 & 70.6 & \textbf{66.1} & - \\
\ + RLSD~\cite{yang2026self}
& \textbf{64.8} & 77.4 & 64.8 & 70.6 & \textbf{45.5} & 54.6 & 71.3 & 79.9 & 63.3 & 63.6 \\
\ + \textbf{AMR-SD (Ours)}
& 59.7 & \textbf{77.6} & \textbf{65.3} & \textbf{79.7} & \textbf{45.5} & \textbf{66.4} & \textbf{74.0} & \textbf{80.3} & 61.0 & 66.1 \\
\bottomrule
\end{tabular}
\end{table*}

\begin{figure*}[t]
  \includegraphics[width=0.48\linewidth]{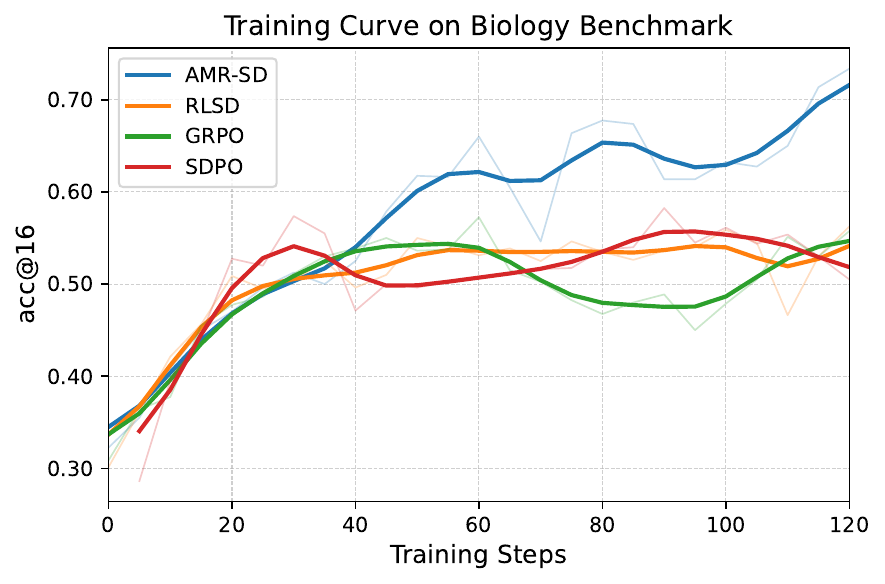} \hfill
  \includegraphics[width=0.48\linewidth]{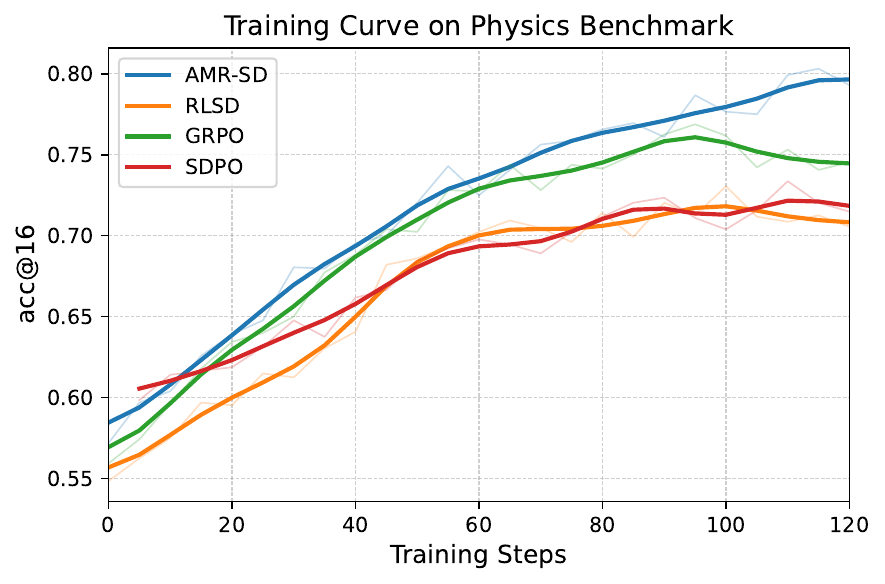}
  \caption{Training dynamics on SciKnowEval Biology (left, Qwen2.5-7B-Instruct) and Physics (right, Qwen3-8B). Unlike baselines that exhibit early gains followed by plateaus or regression, AMR-SD achieves superior stability and avoids late-stage performance degradation across different model architectures.}
  \label{fig:training_curves}
\end{figure*}

\begin{table*}[htbp]
\centering
\caption{Performance of Qwen3-8B on mathematical reasoning benchmarks (Acc@16). AMR-SD consistently outperforms baselines, with significant margins on complex multi-step tasks like AIME24 and HMMT.}
\label{tab:qwen3-math-benchmarks}
\begin{tabular}{l c c c c c}
\toprule
\textbf{Method} & \textbf{AIME24} & \textbf{AIME25} & \textbf{AMC23} & \textbf{HMMT} & \textbf{Avg.} \\
\midrule
\textbf{Qwen3-8B} & 58.1 & 46.5 & 87.5 & 28.3 & 55.1 \\
\ + GRPO~\cite{guo2025deepseek}          & 66.7 & 53.1 & 89.8 & 31.7 & 60.3 \\
\ + RLSD~\cite{yang2026self}          & 63.4 & 47.3 & 90.4 & 27.6 & 57.2 \\
\ + \textbf{AMR-SD (Ours)}           & \textbf{69.6} & \textbf{55.2} & \textbf{91.4} & \textbf{34.6} & \textbf{62.7} \\
\bottomrule
\end{tabular}
\end{table*}

\paragraph{Scientific Reasoning and Tool Use.}
We evaluate AMR-SD on the SciKnowEval benchmark across four scientific disciplines (Chemistry, Physics, Biology, and Materials Science) as well as a Tool Use task, using two backbone models: Qwen3-8B and Qwen2.5-7B-Instruct~(Table~\ref{tab:main_result}). Across both backbones, all training methods substantially improve over the untrained baselines, confirming that reinforcement learning provides a consistent benefit for scientific reasoning. At final convergence (step 75), AMR-SD achieves the strongest or near-strongest performance in the majority of categories. Most significantly, AMR-SD attains the highest Biology scores on both backbones by a substantial margin: 63.7 on Qwen3-8B and 66.4 on Qwen2.5-7B-Instruct, surpassing GRPO by approximately 3.2 and 18.2 percentage points respectively, and exceeding RLSD by roughly 7.2 and 11.8 percentage points. On Physics and Materials Science, AMR-SD similarly reaches competitive or best-in-class scores at step 75---79.7 on Physics and 80.3 on Materials for Qwen2.5-7B-Instruct, the highest figures among all methods in both categories. These results suggest that the token-level credit assignment provided by the CIG mechanism enables more reliable generalisation across diverse scientific domains compared to trajectory-level or coarser reward signals.

\paragraph{Convergence Dynamics.}
A notable pattern across both backbones is the distinct convergence profile of AMR-SD relative to competing methods (Figure~\ref{fig:training_curves}). At the early checkpoint, AMR-SD frequently lags behind baseline methods that aggressively exploit early reward signals. For instance, on the Qwen2.5-7B-Instruct Tool Use task, AMR-SD scores 61.0 at step 15, trailing both GRPO (61.6) and RLSD (63.3). Similarly, on Qwen3-8B Chemistry, AMR-SD (65.0) lags noticeably behind SDPO (69.2) and RLSD (66.3). However, by step 75, these early gaps are closed or reversed, with AMR-SD achieving the best or near-best scores in most categories on both backbones. By contrast, RLSD exhibits early gains that subsequently plateau or regress; its Biology score on Qwen3-8B improves only to 56.5 at step 75---well below AMR-SD's 63.7. SDPO demonstrates the most pronounced instability, suffering severe performance degradation on both backbones and being excluded entirely from the step-75 Tool Use evaluation due to severe policy collapse. These dynamics indicate that AMR-SD's fine-grained credit assignment trades off faster initial convergence for more stable and ultimately superior long-run optimization.

\paragraph{Mathematical Reasoning.}
To assess generalisation beyond scientific knowledge tasks, we further evaluate AMR-SD on four mathematical competition benchmarks---AIME24, AIME25, AMC23, and HMMT---using Qwen3-8B in thinking mode~(Table~\ref{tab:qwen3-math-benchmarks}). AMR-SD achieves the best performance on every individual benchmark and the highest average score of 62.7, compared to 60.3 for GRPO (+2.4 points) and 57.2 for RLSD (+5.5 points). The gains are particularly pronounced on the most challenging benchmarks: 69.6 on AIME24 (vs.\ 66.7 for GRPO), 55.2 on AIME25 (vs.\ 53.1), and 34.6 on HMMT (vs.\ 31.7). Notably, RLSD shows constrained optimization efficiency on harder problems, scoring 27.6 on HMMT and 47.3 on AIME25. This indicates that directly using raw ground-truth solutions as privileged information introduces a distributional shift that makes the resulting reward signal insufficiently discriminative for complex multi-step reasoning. 

Furthermore, we provide deeper analyses of optimization stability and generation dynamics in Appendices \ref{app:training_dynamics} and \ref{app:reflection_analysis}. As shown, while standard continuous self-distillation methods like RLSD exhibit late-stage performance fluctuations and compressed reflection behavior due to direct answer exposure, AMR-SD maintains a stable, monotonically improving reward signal. Crucially, AMR-SD improves reasoning efficiency by reducing unnecessary verbosity while successfully preserving critical cognitive explorations. This confirms that introducing a reflection bottleneck successfully mitigates the vulnerabilities inherent in direct answer conditioning, effectively translating training stability to superior out-of-domain accuracy.

\subsection{Analysis: Quality of Meta-Reflections}

The efficacy of AMR-SD hinges on the quality of self-generated reflections, as they directly calibrate the CIG-based advantage modulation. A precise reflection concentrates the learning signal on semantically informative tokens, whereas a vague or factually incorrect one risks propagating miscalibrated supervision.

Qualitative analysis of sampled reflections from SciKnowEval and DAPO-Math-17k rollouts reveals two consistent patterns. First, \textbf{hints for successful trajectories} effectively distill latent core principles rather than merely repeating the answer path. Second, \textbf{critiques for failed trajectories} demonstrate precise fault localization, diagnosing root causes and reframing them in a positive, prospective manner. This "positive-framing" is crucial; as the critique is prepended to the teacher's forward pass, it provides constructive guidance that enables the model to resolve the underlying reasoning divergence. Detailed representative examples of these Socratic targets are provided in Appendix~\ref{app:reflection_examples}.

\subsection{Ablation Study}

To disentangle the contribution of each component in AMR-SD, we conduct an ablation study on a combined SciKnowEval benchmark, mixing Chemistry, Physics, Biology, and Materials Science for both training and evaluation. All variants are trained on Qwen3-8B and evaluated using Acc@16. Results are reported in Table~\ref{tab:Ablation_Pipeline}.

\begin{table}[htbp]
    \centering
    \caption{Ablation study of the AMR-SD framework on Qwen3-8B (Acc@16). Results validate the necessity of Meta-Reflection, the CIG threshold ($\tau$), asymmetric ReLU gating, and temporal annealing.}
    
    \label{tab:Ablation_Pipeline}
    \resizebox{\linewidth}{!}{
        \begin{tabular}{lc}
            \toprule
            \textbf{Method Setting} & \textbf{Acc@16} \\
            \midrule
            \textit{Qwen3-8B}   & 46.9  \\
            \textit{\textbf{AMR-SD(Full)}}  & \textbf{74.3}  \\
            + \textit{w/o Meta-Reflection}    & 70.5 \\
            + \textit{w/o CIG Threshold ($\tau$)}  & 72.0 \\
            + \textit{w/o CIG Asymmetric ReLU and $\tau$}  & 71.3 \\
            + \textit{w/o Temporal Annealing}  & 70.6 \\
            \bottomrule
        \end{tabular}
    }
\end{table}

\paragraph{Effect of Meta-Reflection.}
In this ablation variant, we replace the model's self-generated meta-reflection with the ground-truth solution directly as the teacher's conditioning context. Performance drops from 74.3 to 70.5, a reduction of 3.8 points. This result reveals a counterintuitive finding: injecting privileged ground-truth information as the conditioning signal is strictly inferior to the model's own Socratic self-reflection. The underlying cause lies in the information asymmetry problem: when the teacher is conditioned on the ground-truth answer, its token probability distribution presupposes information unavailable to the student, creating a fundamental distributional gap that inflates CIG estimates and induces miscalibrated advantage modulation. By contrast, self-generated hints and critiques share the same information boundary as the student policy, yielding better-calibrated CIG signals and more precise gradient interventions.

\paragraph{Effect of the ReLU Threshold $\tau$.}
Removing the margin $\tau$ while retaining the ReLU gate yields 72.0, surpassing the no-reflection baseline but falling 2.3 points below the full model. The threshold enforces sparsity: without it, low-magnitude CIG signals that correspond to distributional noise rather than genuine knowledge deficits are allowed to modulate the advantage, diluting gradient quality. Further ablating the ReLU gate itself reduces accuracy to 71.3. In this setting, the mechanism loses its asymmetric directional filtering. Without the gate, ``counter-directional'' signals---such as positive CIG in a failed rollout or negative CIG in a successful one---can freely offset the intended advantage modulation, partially undermining the primary reward signal. These two ablations validate that gate and margin are necessary for robust, noise-resistant, and directionally-aware credit assignment.

\paragraph{Effect of Temporal Annealing.}
Disabling the decay schedule reduces performance to 70.6. In late training stages, as the student policy becomes highly optimized, its residual errors become increasingly complex. This not only degrades self-reflection quality but also exhausts the teacher model's inherent capacity to provide reliable token-level guidance. Without annealing, persistent distillation forces the policy to learn from these noisy, depleted signals. Temporal annealing mitigates this by smoothly shifting the optimization focus entirely to the pure environmental reward.

\paragraph{Full Model.}
The complete AMR-SD configuration achieves 74.3, demonstrating that all four components are mutually reinforcing rather than individually redundant.

\subsection{Analysis: CIG Token Value Distribution}

As detailed in Appendix~\ref{app:cig_distribution}, the empirical distribution of non-zero CIG values across $n = 50{,}000$ sampled tokens exhibits a pronounced left skew, with the bulk of mass concentrated in the interval $[-1, 0]$ and a peak just below zero. This shape is a direct consequence of the teacher's privileged information. When the teacher model conditions on the meta-reflection, its output distribution is recalibrated, systematically driving the bulk of CIG values below zero.

The asymmetry has a direct consequence for hyperparameter selection. Because tokens satisfying the critique condition vastly outnumber those satisfying the hint condition (as shown in the appendix), setting $\lambda > \gamma$ compensates for this imbalance by up-weighting the sparser positive-CIG tokens. The result is a more balanced effective gradient that leverages both signals without allowing either to overwhelm the other.

\section{Conclusion}

In this paper, we introduced AMR-SD to resolve the token-level credit-assignment bottleneck in RLVR. To overcome the privileged information leakage and signal dampening inherent in standard self-distillation, AMR-SD employs an autonomous Meta-Reflection phase to generate Socratic hints and critiques. We formalized CIG with an asymmetric, ReLU-gated threshold to translate these reflections into sparse, precise token-level interventions, strictly preserving the base environmental reward. Extensive experiments across scientific (SciKnowEval), mathematical (AIME, AMC, HMMT), and tool-use (ToolAlpaca) benchmarks demonstrate that AMR-SD significantly outperforms state-of-the-art baselines. By effectively filtering distributional noise and utilizing temporal annealing, our framework guarantees long-horizon stability and avoids late-stage training collapse, providing a highly robust, scalable solution for aligning large language models on complex reasoning and agentic tasks.

\section*{Limitations}

Despite the strong empirical results, AMR-SD has two notable limitations.

\paragraph{Dependence on Reflection Quality.}
The effectiveness of the CIG mechanism is fundamentally contingent on the quality of the
self-generated meta-reflections. When the base model lacks sufficient introspective capability,
the produced hints and critiques tend to be vague or factually miscalibrated, propagating
corrupted supervision signals through the CIG modulation and ultimately degrading performance.
This manifests most clearly in mathematical reasoning tasks: on weaker or non-thinking-mode
models such as Qwen2.5-7B-Instruct and Qwen3-8B with \texttt{enable\_thinking=False}, the
reflection quality is insufficient to yield reliable CIG signals, and the method fails to
outperform the GRPO baseline. AMR-SD is therefore most effective when deployed on models
with strong chain-of-thought or self-evaluation capabilities.

\paragraph{Inference Overhead from Meta-Reflection Generation.}
The meta-reflection phase introduces an additional autoregressive generation step per trajectory
during training rollouts. Although this overhead is bounded by the reflection budget (capped at
4096 tokens for SciKnowEval and 16384 tokens for math), it nonetheless increases the wall-clock
time per training iteration relative to standard GRPO or RLSD. However, this computational
overhead is highly manageable in practice. Empirically, our method generally maintains a training
speed of at least 80\% (e.g., in the mathematical reasoning task, during the first 50 steps before method degeneration, GRPO and RLSD took 18 hours, whereas AMR-SD took 22 hours) compared to standard GRPO. For severely resource-constrained
settings, this latency may still be a practical concern, and future work could explore
lightweight reflection distillation or early-exit strategies to further reduce this cost.

\bibliography{custom}

\clearpage
\appendix

\section{Prompt Template}
\label{sec:appendix}

\begin{tcolorbox}[breakable, title=Prompt for Generating Hint from Correct Solution]
\small
\texttt{{[}Task{]} \\
You will be provided with a {[}Problem{]} and a {[}Successful Solution{]}. The solution is correct. Your goal is to produce a concise Hint that helps someone solve this type of problem, without revealing the answer. \\
\\
First, write an in-depth {[}Analysis{]} section where you strictly follow these reasoning steps: \\
1. Step-by-step Breakdown: Briefly map out the sequence of logical steps the solution took. \\
2. Verification: Explain WHY these steps successfully satisfy the problem's constraints. \\
3. Detour Check: Identify any points where the solution reconsidered or tried a different approach (if none, explicitly state it was a direct path). \\
4. Core Insight Extraction: Distill the exact mathematical, logical, or factual mechanism that unlocked the answer. \\
\\
Then write your final Hint (2--4 sentences) covering: \\
- A brief summary of the core path (and any real self-corrections, if they occurred). \\
- The key principle or core insight required to solve it. \\
\\
Requirements: \\
1. Do NOT reveal or hint at the final answer. \\
2. Do not invent hesitation or detours if the solution was straightforward. \\
3. If the solution DID explore alternative approaches or self-correct, explicitly preserve and highlight that exploratory path --- this is valuable signal that should NOT be omitted or flattened into a direct narrative. \\
4. Enclose your final Hint strictly within <hint> and </hint> tags. \\
\\
--- \\
{[}Demonstration{]} \\
\\
{[}Problem{]} \\
John has 12 marbles of different colors, including one red, one green, and one blue marble. In how many ways can he choose 4 marbles, if exactly one of the chosen marbles is red, green, or blue? \\
\\
{[}Successful Solution{]} \\
We are given: \\
- John has 12 marbles, each of different colors. \\
- Among them, one red, one green, and one blue marble. \\
- So, there are 3 special marbles (red, green, blue), and the rest are 9 other marbles. \\
- We are to select 4 marbles such that exactly one of the chosen marbles is red, green, or blue. \\
Step 1: Understand the condition --- exactly one must be from \{red, green, blue\}. \\
Step 2: Choose 1 special marble: C(3,1) = 3 \\
Step 3: Choose 3 non-special marbles: C(9,3) = 84 \\
Step 4: Total = 3 $\times$ 84 = 252 \\
\\
{[}Analysis{]} \\
1. Step-by-step Breakdown: The solution categorizes marbles into 3 ``special'' and 9 ``regular'', computes C(3,1) and C(9,3) independently, then multiplies. \\
2. Verification: Separating into mutually exclusive pools enforces the ``exactly one special'' constraint without overcounting. \\
3. Detour Check: No detours --- it was a direct path. \\
4. Core Insight Extraction: When an ``exactly k from group A'' constraint exists, split into independent combination problems on mutually exclusive subsets and apply the multiplication principle. \\
\\
<hint> \\
The solution takes a direct path with no detours. The core insight is mutually exclusive subset division: separate items into a ``target'' group and a ``non-target'' group, compute each selection independently, and multiply. This enforces the ``exactly one'' constraint cleanly without inclusion-exclusion. \\
</hint> \\
\\
--- \\
{[}Actual Task{]} \\
\\
{[}Problem{]} \\
\{prompt\} \\
\\
{[}Successful Solution{]} \\
\{response\} \\
\\
{[}Analysis{]}
}
\end{tcolorbox}

\begin{tcolorbox}[breakable, title=Prompt for Generating Hint from Correct Solution (Thinking Mode)]
\small
\texttt{{[}Problem{]} \\
\{prompt\} \\
\\
{[}Successful Solution{]} \\
\{response\} \\
\\
Carefully analyze the solution step by step, then produce a concise hint (2--4 sentences) that captures the core insight needed to solve this type of problem. Enclose your hint within <hint> and </hint> tags.
}
\end{tcolorbox}

\begin{tcolorbox}[breakable, title=Prompt for Generating Critique from Flawed Solution (with Reference)]
\small
\texttt{{[}Task{]} \\
You will be provided with a {[}Problem{]}, a {[}Reference Correct Solution{]}, a {[}Flawed Solution{]}, and {[}Ground Truth \& Feedback{]}. The flawed solution is incorrect. Your goal is to produce a concise Critique by comparing against the reference. \\
\\
First, write an in-depth {[}Diagnosis{]} section where you strictly follow these reasoning steps: \\
0. Format Check: Check whether the flawed solution's final answer is enclosed in \$\textbackslash boxed\{\}\$. \\
\ \ \ If \$\textbackslash boxed\{\}\$ is MISSING and the mathematical reasoning appears otherwise correct, diagnose \\
\ \ \ this as a FORMAT ERROR --- skip steps 1--3 and write a critique solely about the formatting \\
\ \ \ requirement. If \$\textbackslash boxed\{\}\$ is present, proceed to steps 1--3. \\
1. Reference Anchor: Briefly summarize the key steps and principle in the reference correct solution. \\
2. The Fault Line: Identify the FIRST step in the flawed solution where it diverges from the reference's approach. State what the flawed solution did vs. what the reference did. \\
3. Root Cause Analysis: Identify the underlying misconception or missed constraint that caused the divergence. \\
\ \ \ If no clear error can be identified, leave this step blank. \\
\\
Then write your final Critique (1--2 sentences) stating the correct principle in positive framing --- as if giving a reminder BEFORE someone attempts the problem. Do NOT describe or repeat the error; state only what the reference solution did correctly and what principle must be applied. \\
\\
Requirements: \\
1. POSITIVE FRAMING ONLY: State what should be done correctly, not what went wrong. The critique will be read before the solution, so it must guide toward the right approach. \\
2. BE SPECIFIC: Reference the actual scenario/variables in the problem, not generic rules. \\
3. Enclose your final Critique strictly within \textless critique\textgreater\ and \textless/critique\textgreater\ tags. \\
\\
--- \\
{[}Demonstration{]} \\
\\
{[}Problem{]} \\
Halfway through a 100-shot archery tournament, Chelsea leads by 50 points. Chelsea always scores at least 4 points. If Chelsea's next $n$ shots are bullseyes, she is guaranteed to win. Find the minimum $n$. \\
\\
{[}Reference Correct Solution{]} \\
Chelsea's minimum future score: $10n + 4(50{-}n) = 200 + 6n$. \\
Opponent's maximum future score: $10 \times 50 = 500$. \\
Chelsea's current lead: 50. Guarantee win: (Chelsea's total) $>$ (Opponent's total). \\
Full: $(C_{\text{old}} + 200 + 6n) > (O_{\text{old}} + 500) \Rightarrow 50 + 200 + 6n > 500 \Rightarrow 6n > 250 \Rightarrow n \geq 42$. \\
\\
{[}Flawed Solution{]} \\
Opponent worst-case: $O + 10(50 - n)$. Solving... $n = 16$. \\
\\
{[}Ground Truth \& Feedback{]} \\
Expected Ground Truth: 42 \\
System Feedback: Wrong answer \\
\\
{[}Diagnosis{]} \\
1. Reference Anchor: The reference correctly assumes the opponent takes all 50 remaining shots at maximum (10 pts each = 500), modeling both players' full remaining shot counts independently. \\
2. The Fault Line: The flawed solution uses $10(50{-}n)$ for the opponent's maximum future score, giving the opponent only $50{-}n$ shots. The reference uses $10{\times}50$ because the opponent's shot count is independent of Chelsea's. \\
3. Root Cause Analysis: The flawed solution conflated the two players' remaining shot counts --- Chelsea's $n$ bullseye rounds don't reduce the opponent's turns. \\
\\
\textless critique\textgreater \\
In a guaranteed-victory analysis, each player's remaining shot count must be treated as fully independent. Always model the opponent's worst-case score using their own complete remaining shots, regardless of what Chelsea does in her turns. \\
\textless/critique\textgreater \\
\\
--- \\
{[}Actual Task{]} \\
\\
{[}Problem{]} \\
\{prompt\} \\
\\
{[}Reference Correct Solution{]} \\
\{reference\_response\} \\
\\
{[}Flawed Solution{]} \\
\{response\} \\
\\
{[}Ground Truth \& Feedback{]} \\
Expected Ground Truth: \{ground\_truth\} \\
System Feedback: \{feedback\} \\
\\
{[}Diagnosis{]}
}
\end{tcolorbox}

\begin{tcolorbox}[breakable, title={Prompt for Generating Critique from Flawed Solution (Thinking Mode)}]
\small
\texttt{{[}Problem{]} \\
\{prompt\} \\
\\
{[}Reference Correct Solution{]} \\
\{reference\_response\} \\
\\
{[}Flawed Solution{]} \\
\{response\} \\
\\
{[}Ground Truth \& Feedback{]} \\
Expected Ground Truth: \{ground\_truth\} \\
System Feedback: \{feedback\} \\
\\
Carefully compare the two solutions to identify where the flawed solution diverges, then state the correct principle in 1--2 sentences of positive framing. Enclose your critique within \textless critique\textgreater\ and \textless/critique\textgreater\ tags.
}
\end{tcolorbox}

\section{Reflection Sample}
\label{app:reflection_examples}
\begin{tcolorbox}[breakable, title={Sample: Critique Generation for math(Thinking Mode)}]
\small
\texttt{<think>\\
Okay, let me try to figure out where the flawed solution went wrong. The problem gives us three equations involving a, b, c and x, y, z. The correct approach was to set up a system of equations and then find the determinant of the coefficient matrix, which must be zero for a non-trivial solution. Then, using the given values for x + y + z and xy + yz + zx, we substitute into the determinant equation to solve for xyz.\\
\\
Looking at the correct solution, the determinant expansion led to an expression involving xyz - 2(xy + yz + zx) + 3(x + y + z) - 4 = 0. Substituting the given values, they got xyz = -4044.\\
\\
In the flawed solution, there was a similar determinant setup, but after expanding, they arrived at an expression that included xyz - 2(xy + xz + yz) + 3(x + y + z) - 10 = 0. Solving this gave xyz = -4038. Comparing the two, the key difference is in the constant term in the determinant equation. The correct solution had -4, while the flawed one used -10.\\
\\
I think the mistake in the flawed solution came from an error in simplifying the determinant expansion. Specifically, when expanding the terms, they might have miscalculated the constants or the coefficients. The correct approach carefully expands each term and ensures that all constants are accounted for properly, leading to the correct constant term of -4. This small discrepancy in the constant term led to a different final result.\\
\\
The correct principle here is to carefully expand the determinant, ensuring that all terms are correctly simplified, especially the constants, to match the given conditions. This attention to detail in algebraic manipulation is crucial for arriving at the right answer.\\
</think>\\
\\
<critique>\\
The flawed solution contains an error in simplifying the determinant expression, resulting in an incorrect constant term. This led to a different final value for $xyz$. The correct approach requires careful expansion and careful attention to detail during the algebraic manipulation.\\
</critique>}
\end{tcolorbox}

\begin{tcolorbox}[breakable, title={Sample: Hint Generation for math(Thinking Mode)}]
\small
\texttt{<think>\\
Okay, let's try to figure out how to solve this problem. So, we have a 30-60-90 triangle with a hypotenuse of 20. Three circles are centered at each vertex, and they're all externally tangent to each other. We need to find the area of the union of these circles intersected with the triangle, which is given in the form (m + n$\sqrt{3}$)$\pi$, and then find m + n.\\
\\
First, I remember that in a 30-60-90 triangle, the sides are in the ratio 1 : $\sqrt{3}$ : 2. Since the hypotenuse is 20, the shorter leg (opposite 30°) should be half of that, so 10. The longer leg (opposite 60°) would be 10$\sqrt{3}$. So the triangle has sides 10, 10$\sqrt{3}$, and 20. That makes sense.\\
\\
Now, the circles are centered at each vertex and are externally tangent to each other. That means the distance between any two centers (which are the vertices of the triangle) should be equal to the sum of their radii. So, if we let r\_A, r\_B, r\_C be the radii at vertices A, B, C respectively, we can set up equations based on the sides of the triangle. For example, AB = r\_A + r\_B, AC = r\_A + r\_C, and BC = r\_B + r\_C. Since AB is the hypotenuse, which is 20, AC is 10$\sqrt{3}$, and BC is 10.\\
\\
So we have three equations:\\
1. r\_A + r\_B = 20\\
2. r\_A + r\_C = 10$\sqrt{3}$\\
3. r\_B + r\_C = 10\\
\\
Solving these equations should give the values of the radii. Let me try that. Subtract equation 3 from equation 1: (r\_A + r\_B) - (r\_B + r\_C) = 20 - 10 $\rightarrow$ r\_A - r\_C = 10. So r\_A = r\_C + 10. Then substitute into equation 2: (r\_C + 10) + r\_C = 10$\sqrt{3}$ $\rightarrow$ 2r\_C + 10 = 10$\sqrt{3}$ $\rightarrow$ 2r\_C = 10($\sqrt{3}$ - 1) $\rightarrow$ r\_C = 5($\sqrt{3}$ - 1). Then r\_A = 5($\sqrt{3}$ - 1) + 10 = 5$\sqrt{3}$ + 5, and r\_B = 10 - r\_C = 10 - 5($\sqrt{3}$ - 1) = 15 - 5$\sqrt{3}$. So the radii are found.\\
\\
Now, the next part is to find the area of the union of the circles intersected with the triangle. Since the circles are externally tangent, they don't overlap inside the triangle. Therefore, the area we want is simply the sum of the sectors of each circle that lie within the triangle. Each sector corresponds to the angle at each vertex of the triangle. For example, at vertex A, which is 30°, the sector area would be (30/360)$\pi$r\_A² = (1/12)$\pi$r\_A². Similarly for the other angles.\\
\\
So, compute each sector area:\\
- For angle A (30°): (1/12)$\pi$*(5 + 5$\sqrt{3}$)²\\
- For angle B (60°): (1/6)$\pi$*(15 - 5$\sqrt{3}$)²\\
- For angle C (90°): (1/4)$\pi$*(5$\sqrt{3}$ - 5)²\\
\\
Then add them up. r\_A² = (5 + 5$\sqrt{3}$)² = 25 + 50$\sqrt{3}$ + 75 = 100 + 50$\sqrt{3}$. r\_B² = (15 - 5$\sqrt{3}$)² = 225 - 150$\sqrt{3}$ + 75 = 300 - 150$\sqrt{3}$. r\_C² = (5$\sqrt{3}$ - 5)² = 75 - 50$\sqrt{3}$ + 25 = 100 - 50$\sqrt{3}$.\\
\\
Sector A: (1/12)(100 + 50$\sqrt{3}$)$\pi$ = 25/3$\pi$ + 25$\sqrt{3}$/6$\pi$\\
Sector B: (1/6)(300 - 150$\sqrt{3}$)$\pi$ = 50$\pi$ - 25$\sqrt{3}$$\pi$\\
Sector C: (1/4)(100 - 50$\sqrt{3}$)$\pi$ = 25$\pi$ - 25$\sqrt{3}$/2$\pi$\\
\\
Adding $\pi$ terms: 25/3 + 50 + 25 = 250/3$\pi$. Adding $\sqrt{3}$ terms: 25$\sqrt{3}$/6 - 25$\sqrt{3}$ - 25$\sqrt{3}$/2. Convert to sixths: 25/6 - 150/6 - 75/6 = -200/6 = -100/3. So total area is (250/3 - 100/3$\sqrt{3}$)$\pi$. Therefore m = 250/3, n = -100/3, so m + n = 150/3 = 50.\\
\\
So the key steps were setting up the equations for the radii based on the triangle sides and the external tangency, calculating each sector area based on the angles of the triangle, summing them up, and simplifying. The important thing is that the circles don't overlap inside the triangle, so their sectors can be summed directly.\\
</think>\\
\\
<hint>\\
Use the properties of a 30-60-90 triangle to determine side lengths, then set up a system of equations to find the radii of the externally tangent circles. Calculate the area of the sectors within the triangle, using the triangle's angles, and sum them to find the final area in the form $(m + n\sqrt{3})\pi$, then compute $m + n$.\\
</hint>}
\end{tcolorbox}

\begin{tcolorbox}[breakable, title={Sample: Critique Generation}]
\small
\texttt{[Diagnosis]\\
1. Reference Anchor: The reference solution correctly considers the different types of biological interactions (catalysis, reaction, activation, and expression) and logically narrows down to the most plausible relationship, which is activation given the options.\\
2. The Fault Line: The flawed solution incorrectly chose `Expression' as the relationship, which is a broader term and not as specific as `Activation'.\\
3. Root Cause Analysis: The flawed solution missed the nuance that `Activation' is a more direct regulatory interaction compared to `Expression', which pertains to the regulation of gene production rather than functional interaction.\\
\\
<critique>\\
When determining the specific relationship between two proteins, prioritize direct functional interactions over broader regulatory effects. For the given options, choose a relationship that best describes a specific regulatory or catalytic role rather than a general expression regulation.\\
</critique>}
\end{tcolorbox}

\begin{tcolorbox}[breakable, title={Sample: Hint Generation}]
\small
\texttt{[Analysis]\\
1. Step-by-step Breakdown: The solution examines the amino acid sequence for its composition and potential interactions, then uses logical deduction to identify the most plausible stability score based on the given options and the sequence's characteristics.\\
\\
2. Verification: The reasoning aligns with the typical features of protein folding stability, where a moderate positive score indicates reasonable stability, and the sequence contains both hydrophobic and charged residues, which can contribute to a moderate stability.\\
\\
3. Detour Check: No significant detours or alternative approaches were explored. The reasoning was direct and grounded in the properties of the sequence.\\
\\
4. Core Insight Extraction: The key insight is that the stability score of a protein sequence can be estimated based on its amino acid composition and the nature of the interactions between residues.\\
\\
<hint>\\
The solution directly analyzes the amino acid sequence to estimate stability. The key insight is that a moderate positive stability score is consistent with a sequence containing both hydrophobic and charged residues. This allows for a reasonable balance between stability and flexibility.\\
</hint>}
\end{tcolorbox}

\section{Training Dynamics on Mathematical Reasoning}
\label{app:training_dynamics}

\begin{figure*}[htbp]

  \includegraphics[width=0.48\linewidth]{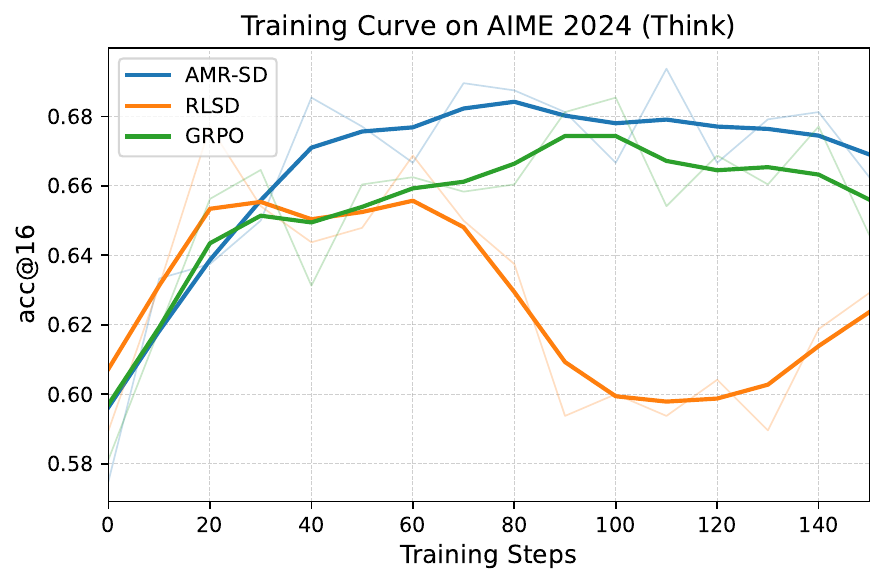} \hfill
  \includegraphics[width=0.48\linewidth]{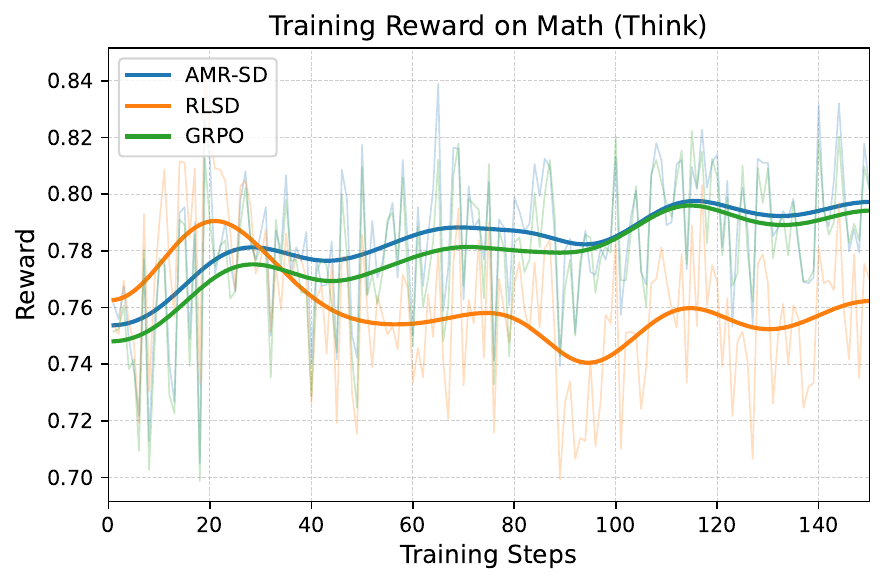}
  \caption{Training dynamics using Qwen3-8B in thinking mode. \textbf{Left:} Evaluation accuracy (acc@16) on the AIME 2024 benchmark over 150 training steps. AMR-SD maintains stable, state-of-the-art performance, whereas RLSD suffers from a severe performance collapse post-annealing. \textbf{Right:} Average training reward on the DAPO-Math-17k dataset. The evaluation collapse of RLSD is accompanied by a sharp degradation in training reward, indicating an intrinsic optimization failure. AMR-SD, by contrast, maintains stable reward growth, successfully bridging in-domain optimization with out-of-domain generalization.}
  \label{fig:aime_training_dynamics}
\end{figure*}

In this section, we provide a detailed visualization of the training dynamics using the Qwen3-8B model with thinking mode enabled. Figure~\ref{fig:aime_training_dynamics} tracks the evaluation accuracy (acc@16) on the out-of-domain AIME 2024 benchmark alongside the average training reward on the DAPO-Math-17k dataset over 150 training steps.

We observe distinct convergence profiles among the evaluated methods. While GRPO demonstrates stable optimization in training reward, its evaluation accuracy quickly plateaus, indicating a generalization bottleneck. Conversely, RLSD experiences a brief initial surge but suffers a severe performance collapse precisely as the distillation signal anneals and diminishes (after step 50). Crucially, the right panel reveals that this evaluation collapse is tightly correlated with a sharp degradation in training reward. This indicates that RLSD's continuous scaling mechanism during the early stages fails to filter out distributional noise, injecting uncalibrated signals that drive the policy into a brittle, overfitted state. Once the distillation guidance is fully decayed, this corrupted policy is unable to sustain optimization under the pure environmental reward, leading to a severe optimization collapse.

In contrast, AMR-SD (aided by the CIG threshold and temporal annealing) achieves a highly stable and continuously improving training reward. More importantly, it successfully translates this internal optimization stability into the highest peak evaluation accuracy on AIME 2024. This confirms that our fine-grained, token-level credit assignment not only stabilizes the RL training process but also significantly enhances the model's true generalization capacity for complex mathematical reasoning.

\section{Hyperparameters and Implementation Details}
\label{app:hyper}
\paragraph{Datasets and Tasks.}
We evaluate AMR-SD on two complementary tracks. For scientific knowledge reasoning and agentic capabilities, we adopt the SciKnowEval benchmark~\cite{feng2024sciknoweval}, spanning four domains (Chemistry, Physics, Biology, and Materials Science), alongside a dedicated Tool Use benchmark~\cite{tang2023toolalpaca}. We follow exactly the training and test splits used in SDPO, enabling a direct and fair comparison. For mathematical reasoning, we train on the preprocessed DAPO-Math-17k dataset~\cite{yu2026dapo} and evaluate on four standard benchmarks: AIME 2024, AIME 2025, AMC 2023, and HMMT.

\paragraph{Base Models and Baselines.}
We conduct experiments on two instruction-tuned base models: \textbf{Qwen3-8B} and \textbf{Qwen2.5-7B-Instruct}~\cite{yang2025qwen3,bai2023qwen}. For all SciKnowEval and Tool Use experiments, we evaluate on both models with the thinking mode disabled (\texttt{enable\_thinking=False}) to ensure a controlled comparison. Conversely, for mathematical reasoning, we exclusively employ \textbf{Qwen3-8B} with the thinking mode enabled (\texttt{enable\_thinking=True}). We compare AMR-SD against the following baselines: (1) the pretrained base model without any RL fine-tuning; (2) \textbf{GRPO}, the standard sequence-level policy gradient baseline; (3) \textbf{SDPO}, a representative on-policy self-distillation method; and (4) \textbf{RLSD}, a state-of-the-art reward-anchored magnitude-adjustment approach. Notably, SDPO is excluded from the mathematical reasoning evaluation due to its severe performance degradation in this domain.

\paragraph{Training Details.}
All models are trained using the \textit{verl} framework~\cite{sheng2024hybridflow} with FSDP distributed training and vLLM-accelerated rollout generation. Regarding task allocation and hardware, the mathematical reasoning model is trained on the preprocessed DAPO-Math-17k dataset. In contrast, the four domains of SciKnowEval and the Tool Use task are trained \textit{independently} for each specific task. For each training prompt, we sample $G=8$ independent trajectories at temperature $1.0$ with top-$p=1.0$. The actor is optimized with AdamW at a learning rate of $1 \times 10^{-6}$, with a training batch size of 256 for SciKnowEval and Tool Use, and 512 for math. The maximum prompt and response lengths are set to 1024 and 2048 tokens respectively for SciKnowEval and Tool Use, and 2048 and 16384 tokens for math. Following the training configuration of RLSD, we set the temporal annealing horizon $T_{\mathrm{decay}} = 50$ steps, after which the objective degenerates smoothly to standard GRPO. The meta-reflection generation budget is capped at 4096 tokens for SciKnowEval and 16384 tokens for math.

Crucially, for the Socratic evaluator $\pi_{\theta_{\text{sg}}}$, we employ a direct evaluation strategy rather than a lagging moving average. During training, the evaluation weights $\theta_{\text{sg}}$ are strictly synchronized with the active policy $\theta$ at every step ($\theta_{\text{sg}} = \text{stop\_gradient}(\theta)$). This design eliminates the need for additional hyperparameter tuning (e.g., EMA decay rates) while ensuring the evaluator remains perfectly coupled with the student's progression. Operating solely in inference mode without gradient tracking, $\pi_{\theta_{\text{sg}}}$ provides a stable, zero-lag baseline for CIG estimation.

\paragraph{Hyperparameter Configurations.}
To account for the highly skewed empirical distribution of CIG values (as detailed in Appendix \ref{app:cig_distribution}), we employ asymmetric modulation coefficients. Specifically, the reward amplification coefficient is set to $\lambda = 0.2$, while the penalty amplification coefficient is set to $\gamma = 0.1$. This configuration effectively up-weights the sparser positive-CIG tokens, ensuring a balanced gradient signal that leverages both hint bonuses and critique penalties.

\paragraph{Evaluation Protocol.}
During evaluation, we strictly control the generation boundaries to match the distinct requirements of each track. For SciKnowEval and the Tool Use benchmark, we set the maximum generation length to 2048 tokens. We report per-domain accuracy at two training checkpoints---step 15 and step 75---to assess both early convergence and final performance, with validation performed every 5 training steps. For mathematical reasoning, we expand the maximum generation length to 16384 tokens to accommodate extended and complex reasoning chains. Performance is evaluated using acc@16, computed as the average accuracy over 16 independently sampled rollouts per problem at temperature $1.0$, across AIME 2024, AIME 2025, AMC 2023, and HMMT.

\section{Detailed Analysis: CIG Token Value Distribution}
\label{app:cig_distribution}

This section provides the visual empirical distribution and a detailed theoretical analysis of the Causal Information Gain (CIG) values discussed in Section 4.5 of the main text.

\begin{figure}[htbp]
  \centering
  \includegraphics[width=0.6\columnwidth]{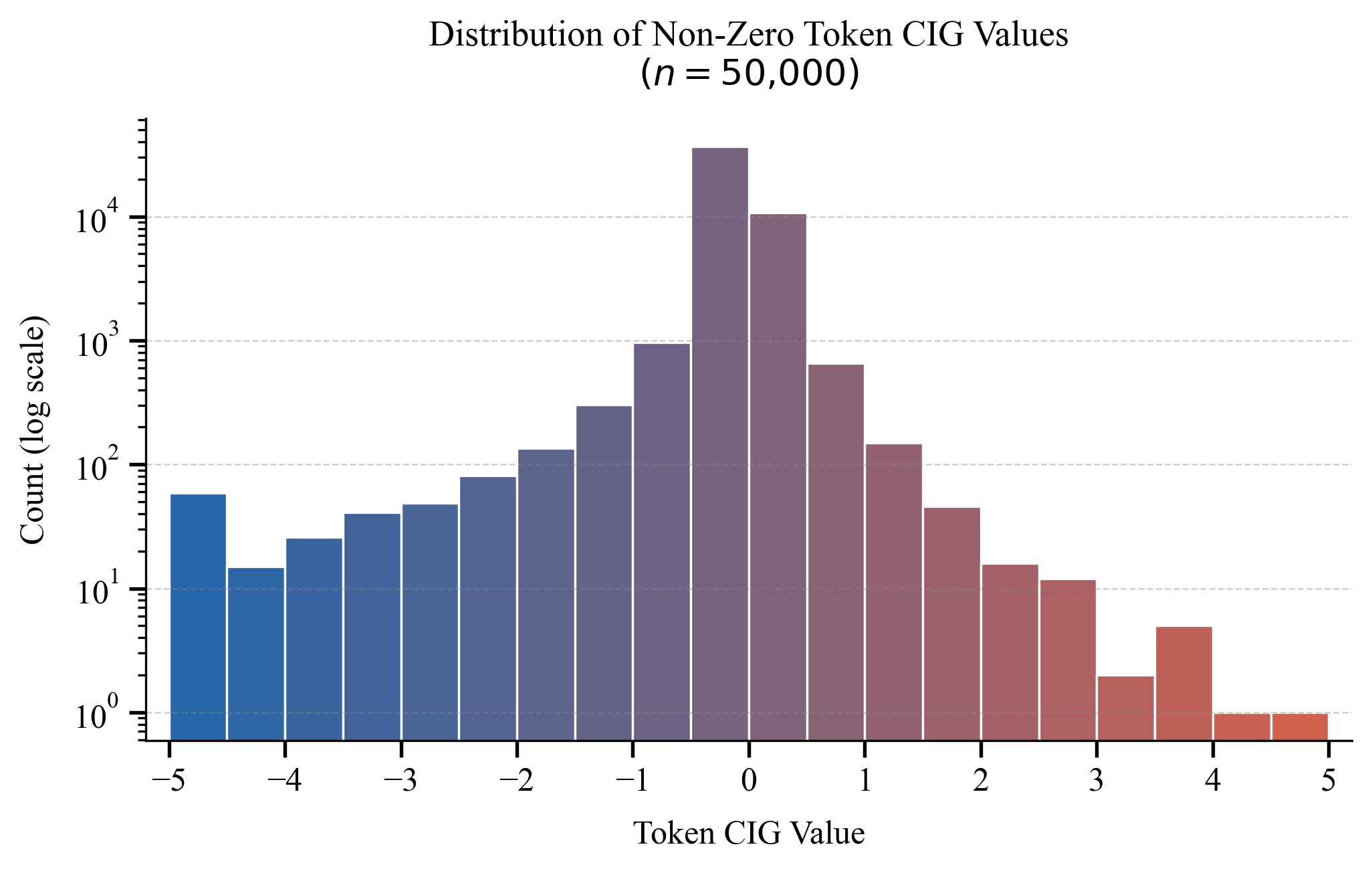}
  \caption{Empirical distribution of non-zero CIG values. The prominent left skew (peaking just below zero) is caused by the teacher's conditioning on privileged information. This intrinsic asymmetry necessitates differential scaling coefficients ($\lambda > \gamma$) to balance penalties and bonuses.}
  \label{Analysis:cig}
\end{figure}

As presented in Figure~\ref{Analysis:cig}, the empirical distribution of non-zero CIG values across $n = 50{,}000$ sampled tokens exhibits a pronounced left skew, with the bulk of mass concentrated in the interval $[-1, 0]$ and a peak just below zero. This shape is a direct consequence of the teacher's privileged information. 

When the teacher model conditions on the meta-reflection---a hint or critique unavailable to the student---its output distribution is recalibrated: probability mass shifts toward alternative continuations, and by the softmax normalization constraint, the probability assigned to the token actually produced by the student decreases. This effect is systematic and occurs across both correct and incorrect trajectories, driving the bulk of CIG values below zero.

The asymmetry has a direct consequence for hyperparameter selection. Recall that the hint bonus activates on correct trajectories when $\mathrm{CIG}_t > \tau$, while the critique penalty activates on wrong trajectories when $-\mathrm{CIG}_t > \tau$. As Figure~\ref{Analysis:cig} shows, the negative tail is substantially heavier: tokens satisfying the critique condition vastly outnumber those satisfying the hint condition. If $\lambda = \gamma$, the aggregate gradient signal would be dominated by critique penalties, suppressing the learning contribution from hint bonuses. Setting $\lambda > \gamma$ compensates for this imbalance by up-weighting the sparser positive-CIG tokens. The result is a more balanced effective gradient that leverages both the reinforcement of teacher-preferred tokens on correct trajectories and the attenuation of teacher-opposed tokens on wrong ones, without allowing either signal to overwhelm the other.

\section{Analysis of Reflection Capabilities and Reasoning Efficiency}
\label{app:reflection_analysis}

\paragraph{Motivation.} 
While our Asymmetric Meta-Reflective Self-Distillation (AMR-SD) framework significantly improves accuracy and reasoning efficiency, a critical concern in On-Policy Distillation is whether the dense supervision alters the model's intrinsic ability to explore and reflect. Standard self-distillation often forces the student to mimic an oracle-conditioned teacher. When this teacher is directly exposed to ground-truth answers, it can induce over-alignment and restrict natural trial-and-error, resulting in more short-sighted generation. To investigate this, we analyze the generation dynamics of different alignment strategies by quantifying their reflection capabilities.

\paragraph{Methodology.}
We sample $n=480$ evaluation traces from the AIME 24 benchmark. To measure the degree of internal deliberation, we track the frequency of ten explicit reflection and self-correction markers (e.g., \textit{Wait, Maybe, Check, Alternatively, Seems, Might, Hmm, Perhaps, Actually, Likely}, following \citealp{kim2026does}) across four distinct models: the Base model (Qwen3-8B), standard GRPO, RLSD, and our proposed AMR-SD.

\begin{table}[h]
\centering
\small
\caption{Frequency of key reflection and self-correction markers across 480 AIME 24 evaluation traces. AMR-SD successfully reduces the excessive verbosity of the base model while preserving the core reflection capabilities that are otherwise compressed in direct answer-conditioned self-distillation.}
\label{tab:reflection_markers}
\resizebox{\columnwidth}{!}{%
\begin{tabular}{l c c}
\toprule
\textbf{Method} & \textbf{Total Count} & \textbf{Avg. per Trace} \\
\midrule
\textbf{Base Model} (Qwen3-8B) & 67,632 & 140.90 \\
\textbf{GRPO} \cite{guo2025deepseek} & 56,432 & 117.57 \\
\textbf{RLSD} \cite{yang2026self} & 35,324 & 73.59 \\
\midrule
\textbf{AMR-SD} (Ours) & 50,553 & 105.32 \\
\bottomrule
\end{tabular}%
}
\end{table}

\paragraph{Results and Discussion.}
As shown in Table~\ref{tab:reflection_markers}, the Base model exhibits extreme verbosity (140.90 markers per trace), often over-thinking without converging to a solution. While GRPO reduces some redundancy (117.57), it still relies heavily on prolonged generation to maximize rewards.

Crucially, standard RLSD experiences a shift in reflection markers, dropping to an average of 73.59 per trace. This strongly confirms that previous self-distillation methods—which directly expose the ground-truth answer as privileged information—tend to over-condition the teacher's distribution. Because the teacher already contains the terminal answer explicitly, its rescored probability landscape implicitly disincentivizes the student's natural logical branching or explicit trial-and-error attempts. Consequently, the model shifts towards more accelerated, direct paths, compressing its long-horizon exploration.

In contrast, our AMR-SD framework strikes an optimal balance, maintaining a healthy reflection frequency of 105.32 markers per trace. This "Goldilocks effect" is directly attributable to replacing direct answer exposure with an information bottleneck (Meta-Reflection) combined with the Asymmetric Causal Information Gain (CIG) threshold. Rather than penalizing cognitive explorations, the asymmetric threshold acts as a precise filter: it heavily penalizes blindly confident deviations (baseless waffle) while tolerating and protecting valid analytical trajectories (e.g., \textit{``Wait, alternatively...''}). This demonstrates that AMR-SD effectively enhances reasoning efficiency without stripping the model of the crucial critical thinking capabilities required for solving complex mathematical problems.

\end{document}